# Assessing Car Damage using Mask R-CNN

Sarath P, Soorya M, Shaik Abdul Rahman A, S Suresh Kumar, K Devaki

*Abstract: Picture based vehicle protection handling is a significant region with enormous degree for mechanization. In this paper we consider the issue of vehicle harm characterization, where a portion of the classifications can be fine-granular. We investigate profound learning based procedures for this reason. At first, we attempt legitimately preparing a CNN. In any case, because of little arrangement of marked information, it doesn't function admirably. At that point, we investigate the impact of space explicit pre-preparing followed by tweaking. At last, we explore different avenues regarding move learning and outfit learning. Trial results show that move learning works superior to space explicit tweaking. We accomplish precision of 89.5% with blend of move and gathering learning.*

## I. INTRODUCTION

Today, in the vehicle insurance industry, a lot of money is wasted due to cases spillage [1] [2]. Cases spillage/Underwriting spillage is described as the differentiation between the genuine case portion made and the entirety that should have been paid if all business driving practices were applied. Visual assessment and endorsement have been used to reduce such effects. In any case, they present delays for the situation planning. There have been tries by a few new organizations to direct case taking care of time [3] [4]. A mechanized structure for the vehicle security ensure planning is a need critical.

In this paper, we use Convolutional Neural Network (CNN) based systems for plan of vehicle hurt sorts. Specifically, we consider typical damage types, for instance, watch gouge, passage engrave, glass break, head light broken, tail light broken, scratch and smash. To the extent we might know, there is no straightforwardly open dataset for vehicle hurt gathering. Thusly, we made our own special dataset by social event pictures from web and physically explaining them. The request task is attempting a direct result of factors, for instance, huge between class resemblance and hardly unquestionable damages. We investigated various roads with respect to various techniques, for instance, clearly setting up a CNN, pre-setting up a CNN using auto-encoder followed by changing, using move picking up from gigantic CNNs arranged on ImageNet and building a gathering classifier over the course of action of pre-arranged classifiers. We see that move learning got together with outfit learning works the best. We moreover devise a method to confine a particular mischief type. Test outcomes affirm the feasibility of our proposed game plan.

**Sarath P.,** Student, Dept of CSE, Rajalakshmi Engineering College, Chennai, TN, India.

**Soorya M.,** Student, Dept of CSE, Rajalakshmi Engineering College, Chennai, TN, India.

**Shaik Abdul Rahman A.,** Student, Dept of CSE, Rajalakshmi Engineering College, Chennai, TN, India.

**S. Suresh Kumar,** Associate Professor, Dept of CSE, Rajalakshmi Engineering College, Chennai, TN, India.

**K Devaki,** Professor in the Department of Computer Science and Engineering, Rajalakshmi Engineering College , Thandalam, Chennai, India.

## II. RELATED WORKS

Significant learning has shown promising results in AI applications. In particular, CNNs perform well for PC vision tasks, for instance, visual thing affirmation and acknowledgment [5] [6]. Utilization of CNNs to fundamental mischief assessment has been mulled over in [7]. The designers propose a significant learning based methodology for Structural Health Monitoring (SHM) to depict the mischief as breaks on a composite material. Solo depiction is used and conclusion are showed up on a broad extent of stacking actions with foreordained number of checked getting ready picture data. Most of the oversaw systems need a ton of checked data and figure resources. Solo pre-getting ready frameworks, for instance, Autoencoders [8] have been shown to im-exhibit the theory execution of the classifier if there ought to be an event of humble number of named tests. For pictures, Convolutional AutoEncoders (CAE) [9] have shown promising results.

A very notable strategy which has worked viably if there should be an occurrence of little named information is move learning. A framework which is set up on a origin task is used as a segment extractor for destination task. There are various CNN models arranged on Imagenet that are available openly, for instance, VGG-16 [12], VGG-19 [12], Alexnet [6], Inception [13], Cars [14], Resnet [15]. Transferable part depiction learned by CNN limits the effect of over-fitting if there ought to be an event of somewhat named set.

Customary AI frameworks have also been tried for robotized hurt assessment. Jaywardena et al [16] proposed a procedure for vehicle scratch hurt revelation by selecting 3D CAD model of entire automobile on the image of the harmed automobile. There has been tries to look at harm in land regions using satellite pictures [17] [18] [19]. To best of our knowledge, significant learning based strategies have not been used for electronic vehicle harm gathering, especially for the fine-granular mischief course of action.

## III. DATASET DESCRIPTION

Considering there is no openly accessible dataset for vehicle harm characterization, we made our own dataset comprising of pictures having a place with various kinds of vehicle harm. We consider seven normally watched sorts of harm, for example, guard gouge, entryway imprint, glass break, head light broken, tail light broken, scratch and crush. Moreover, we additionally gathered pictures which have a place with a no harm class. The pictures were gathered from web and were physically commented on. Table I shows the portrayal of the dataset.

### A. Data augmentation

It is realized that an increase of the dataset with relative changed pictures improves the speculation execution of the classifier. Thus, we artificially augmented the dataset

**Assessing Car Damage using Mask R-CNN**

**TABLE I. Characterization of our dataset**

| Classes | Train size | Augm. train size | Test size |
|---|---|---|---|
| Guard Gouge | 172 | 1254 | 47 |
| Enterway Imprint | 145 | 825 | 38 |
| Glass break | 205 | 1270 | 52 |
| Head-light broken | 192 | 1172 | 46 |
| Tail-light broken | 77 | 484 | 23 |
| Scratch | 186 | 1116 | 46 |
| Crush | 192 | 1094 | 44 |
| No damage | 1282 | 7525 | 320 |

Approximately on numerous occasions by attaching it with subjective transformations (be-tween -20 to 20 degrees) and level flip changes. For the portrayal investigates, the dataset was erratically part into 79%-20% where 79% was used for getting ready and 20% was used for testing. Table I delineates the magnitude of our train and test sets.

Fig. 1 shows test pictures for each class. Note that the request task is non-immaterial as a result of high between class resemblance. Especially, since the mischief doesn't cover the entire picture (yet a little territory of it), it renders game plan task much dynamically irksome.

## IV. OUTLINE OF METHODOLOGY

In the principle course of action of preliminaries, we arranged a CNN starting with the sporadic instatement. Our CNN building include 10 layers: Convolution1-Pooling1-Convolution2-Pooling2-Convolution3-Pooling3-Convolution4-Pooling4-FC-Softmax where Convolution, Pooling, FC and Softmax demonstrates convolution layer, pooling layer, totally related layer and a softmax layer exclusively. Each convolutional layer has 16 channels of size 5x5. A RELU non-linearity is used for each convolutional layer. Unquestionably the quantity of burdens in the framework are around 423K. Dropout was added to each layer which is known to improve hypothesis execution. We arranged a CNN on the main similarly as on the expanded dataset.

Table II shows the delayed consequence of the CNN planning from random instatement. It might be seen that the data development to be certain improves the theory and gives better execution that basically planning on the first data-set.

We know that the information utilized for preparing the CNN (significantly after enlargement) is very less contrasted with the quantity of parameters and it might bring about overfitting. In any case, we played out this analysis to set a benchmark for rest of the investigations.

**TABLE II. Test exactness with CNN preparing and (CAE + tweaking).**

| Method | Without Augmentation | | | With Augmentation | | |
|---|---|---|---|---|---|---|
| | Acc | Prec | Recall | Acc | Prec | Recall |
| CNN | 70.33 | 64.26 | 54.4 | 74.46 | 65.03 | 62.01 |
| AE-CNN | 73.43 | 68.21 | 54.32 | 72.30 | 66.69 | 60.48 |

### A. Convolutional Autoencoder

Solo pre-preparing is a notable procedure, known to be valuable in situations where preparing information is scarce[8].

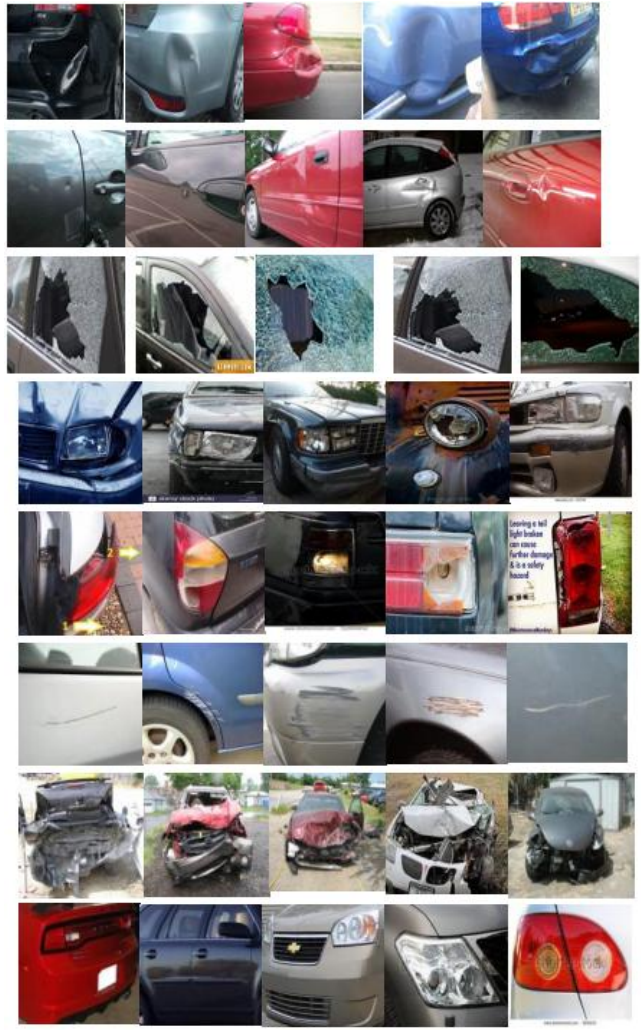

**Fig. 1. Sample pictures for vehicle harm types. Lines through and through shows harm types Bumper imprint, Door mark, Glass break, Head-light broken, Tail-light broken, Scratch, Smash and No harm**

The basic objective of a performance learning procedure is to isolate important features from the unlabeled dataset by learning the data scattering. They perceive and remove input redundancies, and regularly simply protect essential pieces of the data which will by and large assistance the portrayal task. A totally related auto-encoders, especially in case of pictures, prompts gigantic number of trainable parameters. Convolutional AutoEncoders (CAE) give a predominant alternative by virtue of less number of parameters in light of small affiliations and weight sharing [9]. CAEs are set up in a layer sagacious way where solo layers can be stacked more than each other to collect the dynamic framework. Each layer is arranged unreservedly of resulting layer. Finally, the all out course of action of layers are stacked and changed by back-spread using the cross-entropy target work . Independent instatement will when all is said in done keep up a vital good ways from close by minima and increase the frameworks execution robustness.

For preparing a CAE, we utilized unlabeled pictures from Stand ford car dataset [20].

The size of the dataset was artificially expanded by including turn and flip changes. Considering the objective pictures have a place with vehicle harm type, we expect that learning the vehicle explicit highlights should enable the order to task. The layers are then adjusted utilizing a littler learning rate when contrasted with the preparation. The row, AE-CNN, in Table II shows the outcome with autoencoder pre-preparing. It very well may be seen that an autoencoder pre-preparing helps the grouping task. A comparative examination was performed utilizing expanded vehicle harm pictures and there also we consider improvement to be the test precision when contrasted with no pre-preparing.

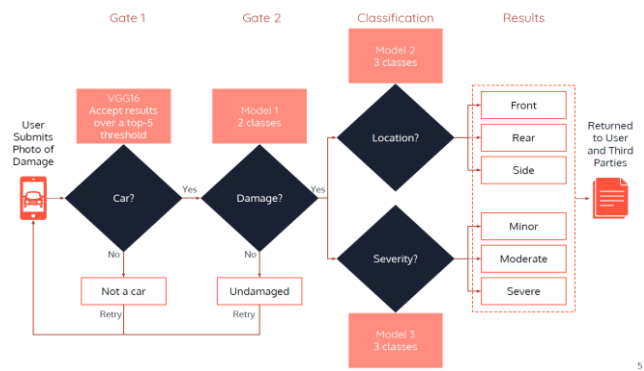

**Fig. 2. Pipeline**

## V. TRANSFER LEARNING

Move learning has indicated promising outcomes if there should arise an occurrence of little marked. In the exchange getting the hang of setting, information from the source task is moved to the objective errand. The instinct is that some information is explicit to singular areas, while some information might be basic between various spaces which may improve execution for the objective space/task. Be that as it may, in the situations where the source space and target area are not identified with one another, savage power move might be fruitless and can prompt the debased presentation. For our circumstance, we use the CNN models which are set up on the Image-net dataset. Since the Image-net dataset contains vehicle as a class, we foresee that the trade should be useful which we broadly affirm by investigating various roads with respect to different pre-arranged models. Fig. 2 shows the exchange learning test arrangement we use. Since we use pre-prepared models which are prepared for ImageNet, the Source task is the ImageNet order. The pre-prepared model is utilized as highlight extractor.

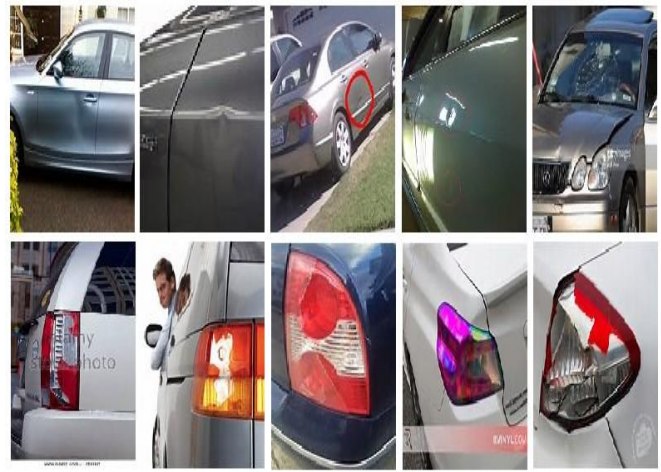

**Fig. 3. Examples of test images mis-classified as 'no damage' class with Resnet. Note that the damaged portion is scarcely visible.**

Table III demonstrates the exactness, accuracy and review when utilizing these pre-prepared models. It very well may be noticed that the Resnet plays out the better amid all the pre-prepared models. The information increase supports the presentation in the greater part of the cases. Midst the analysis, it was seen that the Softmax classifier works superior to direct SVM and it is quicker to prepare. Shockingly, the GoogleNet pre-prepared model calibrated utilizing vehicle dataset, played out the most noticeably terrible. It demonstrates that lone vehicle object based highlights may not be compelling for ordering harm types. The terrible showing of autoencoder based methodology should be because of this impact. It underlines the viability of highlight portrayal gained from huge and different information conveyances.

We see that the central point in the mis-arrangements is the equivocalness among harm and 'no harm' class. This isn't amazing in light of the fact that, the harm of a section as a rule involves an exceptionally little part of the picture and renders recognizable proof troublesome in any event, for the human eyewitness. Fig. 3 shows hardly any instances of test pictures of harm which are mis-named no harm.

### A. Ensemble method

To additionally improve the exactness, we played out an exper-iment with outfit of the pre-prepared classifiers. For each preparation picture, class likelihood expectations are gotten from numerous pre-prepared systems. The weighted

**TABLE III. Order execution for move learning. Examination of test exactnesses with various pre-prepared CNN models. Note that Resnet plays out the best.**

| Model | Params | Dim | Without Augmentation | | | | | | With Augmentation | | | | | |
| --- | --- | --- | --- | --- | --- | --- | --- | --- | --- | --- | --- | --- | --- | --- |
| | | | Definite SVM | | | Softmax | | | Linear SVM | | | Softmax | | |
| | | | Acc | Prec | Recall | Acc | Prec | Recall | Acc | Prec | Recall | Acc | Prec | Recall |
| Cars [14] | 7.5M | 1024 | 58.34 | 57 26 | 57 45 | 65 40 | 57 25 | 35 40 | 59 48 | 50.55 | 56 97 | 68 26 | 55 74 | 40 20 |
| Inception [13] | 6M. | 2048 | 72.12 | 58 46 | 58 53 | 80 85 | 64 78 | 58 72 | 69 70 | 60.60 | 54 47 | 75 57 | 70 48 | 55 85 |
| Alexnet | 72 M. | 4096 | 73.72 | 65 68 | 64 60 | 72 87 | 66 45 | 60 10 | 74 28 | 65.86 | 65 75 | 76 95 | 67 85 | 65 38 |
| VGG-19 [12] | 165M. | 4096 | 94.87 | 79 65 | 75 18 | 88 33 | 84 79 | 74 62 | 85 40 | 80.50 | 80 70 | 85 95 | 85 78 | 75 45 |
| VGG-16 [12] | 142M. | 4096 | 94.74 | 78 79 | 78 42 | 88 88 | 85 95 | 75 57 | 87 95 | 79.65 | 75 98 | 86 78 | 80 99 | 78 36 |
| Resnet [15] | 28.7M. | 2048 | 88.31 | 85 87 | 80 32 | 89.14 | 86 40 | 82 20 | 88 95 | 88.50 | 80 98 | 88 95 | 85 70 | 80 67 |

## B. Experimental Results

**TABLE IV. Order execution for Troupe procedure using Top-3 and full models**

| Ensemble | Without Augmentation | | | With Augmentation | | |
|---|---|---|---|---|---|---|
| | Acc | Prec | Recall | Acc | Prec | Recall |
| Top-3 | 89.39 | 88.12 | 80.94 | 88 30 | 87 88 | 79 93 |
| All | 89.57 | 88.16 | 80.95 | 88 26 | 86 47 | 79 43 |

(Top-3 and full) works superior to anything the individual classifiers, true to form.

## C. Damage localization

With a similar methodology, we can even restrict the harmed part. Now every dot in the analysis picture, we crop an area of size 100x100 over it, change the size to 224x224 and anticipate the class rear ends. A harm is viewed as recognized if the likelihood esteem is over sure limit. Fig. 4 shows the confinement execution for harm types, for example, glass break, crush and scratch with Resnet classifier and likelihood edge of 0.9.

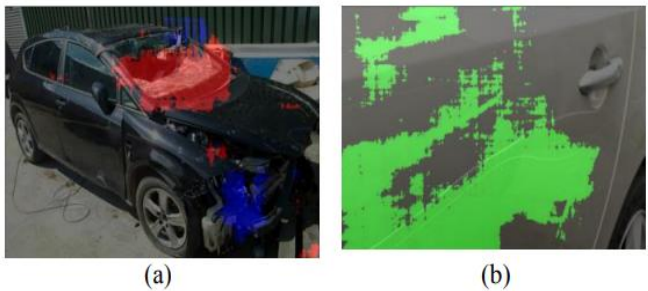

**Fig. 4. Harm confinement. (a) Glass break (red) and Smash (blue), (b) Scratch (green). Note that our methodology can confine harm accurately**

## VI. CONCLUSION

In this paper, we schemed a profound learning based response for vehicle hurt request. Considering there were no straightforwardly available dataset, we made another highlights may not be viable for harm grouping. It therefore underlines the predominance of highlight portrayal gained from the huge preparing set.

## AUTHORS PROFILE

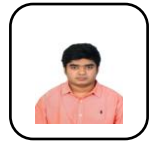

**Sarath P** is a student in the Department of Computer Science and Engineering at Rajalakshmi Engineering College(Anna University). His areas of specialization includes Data mining and Network Security. He is also a Certified Ethical Hacker and also a Pega CSSA.

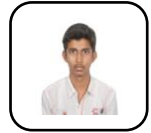

**Soorya M** is a student in the Department of Computer Science and Engineering at Rajalakshmi Engineering College(Anna University). His areas of specialization includes Algorithm Analysis. He is also a Pega CSSA.

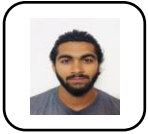

**Shaik Abdul Rahman A** is a student in the Department of Computer Science and Engineering at Rajalakshmi Engineering College(Anna University). His areas of specialization includes Data Mining.

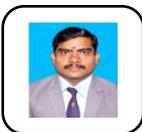

**S. Suresh Kumar** is an Associate Professor in the Department of Computer Science and Engineering, Rajalakshmi Engineering College , Thandalam, Chennai, India. He obtained his Bachelor of Engineering in Computer Science and Engineering from University of Madras, Chennai and Master of Engineering in Computer Science and Engineering from Anna University Chennai .His current research is focused on cloud computing Security.

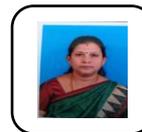

**K Devaki** is a Professor in the Department of Computer Science and Engineering, Rajalakshmi Engineering College , Thandalam, Chennai, India. She obtained her Bachelor of Engineering in Computer Science and Engineering, Master of Engineering and PhD in Computer Science and Engineering.Her current research is focused on Data Mining.